\documentclass{article}

\usepackage[
letterpaper,
left=1in,
right=1in,
top=0.3in,
bottom=1in
]{geometry}

\usepackage{graphicx}
\usepackage{amsmath}
\usepackage{multicol}
\usepackage{hyperref}
\usepackage{cleveref}
\usepackage{lipsum}
\usepackage{abstract}
\usepackage{booktabs}

\usepackage[numbers,sort&compress]{natbib}
\title{
	\noindent\rule{\textwidth}{4pt}
	\textbf{Benchmarking Offline Multi-Objective Reinforcement Learning in Critical Care}
	\noindent\rule{\textwidth}{1pt}
}
\author{\textbf{Aryaman Bansal}$^{1,3}$ \quad \textbf{Divya Sharma}$^{1,2,3}$}
\date{}

\begin{document}
	
	\maketitle
	
	\vspace{-2em} 
	\begin{center}
		University of Toronto$^1$  \quad York University$^2$ \quad University Health Network$^3$ \\
		aryaman.bansal@mail.utoronto.ca \quad div.sharma@utoronto.ca
	\end{center}
	
	\begin{abstract}
		\normalsize
		In critical care settings such as the Intensive Care Unit, clinicians face the complex challenge of balancing conflicting objectives, primarily maximizing patient survival while minimizing resource utilization (e.g., length of stay). Single-objective Reinforcement Learning approaches typically address this by optimizing a fixed scalarized reward function, resulting in rigid policies that fail to adapt to varying clinical priorities. Multi-objective Reinforcement Learning (MORL) offers a solution by learning a set of optimal policies along the Pareto Frontier, allowing for dynamic preference selection at test time. However, applying MORL in healthcare necessitates strict offline learning from historical data.   
		
		In this paper, we benchmark three offline MORL algorithms, Conditioned Conservative Pareto Q-Learning (CPQL), Adaptive CPQL, and a modified Pareto Efficient Decision Agent (PEDA) Decision Transformer (PEDA DT), against three scalarized single-objective baselines (BC, CQL, and DDQN) on the MIMIC-IV dataset. Using Off-Policy Evaluation (OPE) metrics, we demonstrate that PEDA DT algorithm offers superior flexibility compared to static scalarized baselines. Notably, our results extend previous findings on single-objective Decision Transformers in healthcare, confirming that sequence modeling architectures remain robust and effective when scaled to multi-objective conditioned generation. These findings suggest that offline MORL is a promising framework for enabling personalized, adjustable decision-making in critical care without the need for retraining.
	\end{abstract}
	
	\section{Introduction}
	
	In high-stakes domains such as healthcare, Reinforcement Learning (RL) algorithms rarely face a single objective. A clinician treating a patient in the ICU must balance conflicting objectives, primarily maximizing survival probability (minimizing mortality) while minimizing the burden on hospital resources (Length of Stay or LOS). While standard RL simplifies this into a single scalar reward (e.g., a weighted sum), this approach is rigid and fails to capture the dynamic trade-offs required for personalized care.
	For example, suppose mortality and length of stay were used as rewards $R_1$ and $R_2$ with preferences 0.7 and 0.3, respectively. Thus, resulting in the scalarized reward: $R = 0.7 R_1 + 0.3 R_2$. This poses a significant problem because the RL algorithm trained on the preferences 0.7 and 0.3 cannot be used for other preferences, say 0.5 and 0.5, as the RL algorithm was optimized for that. 
	
	Multi-Objective Reinforcement Learning (MORL)~\citep{hayes2021practical} aims to solve this by learning a set of optimal policies along the Pareto Frontier. In other words, it solves this problem by training algorithms that do not optimize for a fixed set of preferences, which in turn allows for any preference to be used at test time. However, applying MORL in healthcare requires an \textit{offline} approach, learning strictly from historical data (e.g., MIMIC-IV~\citep{johnson2023mimic} dataset) without interacting with patients.
	
	In this paper, we will benchmark the performance of 3 MORL algorithms against 3 Single-Objective Reinforcement Learning (SORL) on the MIMIC-IV dataset using 2 off-policy evaluation (OPE) metrics: FQE~\citep{le2019batch} and WIS~\citep{puaduraru2013empirical, voloshin2019empirical}.

	This paper investigates the hypothesis that Offline MORL (OMORL) algorithms should replace scalarized Offline SORL (OSORL) algorithms. We evaluate three MORL candidates: two variants of Conservative Pareto Q-Learning (CPQL)~\citep{van2014multi, bhardwaj2024enhancing, kumar2020conservative} and a modified Pareto Efficient Decision Agent (PEDA)~\citep{zhu2023scaling} based on the Decision Transformer~\citep{chen2021decision}. We compare these against scalarized baselines: Behavior Cloning (BC)~\citep{pomerleau1988alvinn, bain1995framework}, Conservative Q-Learning (CQL)~\citep{kumar2020conservative}, and Double Deep Q-Network (DDQN)~\citep{van2016deep}.
	
	Our contributions are as follows:
	\begin{enumerate}
		\item We propose and evaluate Conditioned CPQL and Adaptive CPQL, comparing them against a modified Transformer-based~\citep{vaswani2017attention} algorithm (PEDA DT).
		\item We demonstrate that OMORL algorithms (specifically PEDA DT) show promise under WIS and FQE.
	\end{enumerate}

	%
	%
	
	\section{Methodology}
	
	%
	
	%
	%
	
	\subsection{Scalarized OSORL Baselines}
	To provide a strong benchmark, we trained three established offline RL algorithms using a fixed scalarization strategy. The reward function for these baselines was defined as:
	\begin{equation}
		R_{scalar} = 0.5 \cdot r_{mortality} + 0.5 \cdot r_{LOS}
	\end{equation}
	The baselines selected are:
	\begin{itemize}
		\item BC: A supervised learning approach that mimics the clinician's policy found in the dataset.
		\item DDQN: A value-based method (uses a neural network to approximate Q-values) adapted for offline learning.
		\item CQL: A state-of-the-art offline method that explicitly penalizes Q-values for out-of-distribution actions to prevent overestimation.
	\end{itemize}
	
	\subsection{OMORL Architectures}
	We evaluate three multi-objective approaches designed to handle the preference vector $\omega$ explicitly. Conditioned CPQL (C-CPQL) uses static conditioning via concatenation. Adaptive CPQL (AP-CPQL) employs a dynamic gating mechanism with a Preference Attention (PA) block. Finally, we employ a Modified PEDA Decision Transformer (PEDA DT), adapted for variable length trajectories and offline evaluation.
	
	\section{Experimental Setup}
	
	\subsection{Dataset}
	Experiments were conducted using the MIMIC-IV dataset. We extracted patient trajectories containing physiological state features (e.g., vitals, lab results) and interventions.
	
	\subsection{Evaluation Protocols}
	We utilized three standard Off-Policy Evaluation (OPE) metrics to assess performance without deploying policies on real patients. These metrics are defined~\citep{rahman2024empowering} as follows:
	
	\subsubsection{Weighted Importance Sampling (WIS)}
	WIS uses a behavior policy $\pi_b$ to evaluate a policy by re-weighting episodes according to their likelihood of occurrence. With the per-step importance ratio $\rho_t = \frac{\pi(a_t|s_t)}{\pi_b(a_t|s_t)}$ and cumulative importance ratio $\rho_{1:t} = \prod_{t'=1}^{t} \rho_{t'}$, WIS is computed as:
	
	\begin{equation}
		V_{WIS}(\pi) = \frac{1}{N} \sum_{n=1}^{N} \frac{\rho_{1:T^{(n)}}^{(n)}}{w_{T^{(n)}}} \left( \sum_{t=1}^{T^{(n)}} \gamma^{t-1} r_t^{(n)} \right)
	\end{equation}
	
	where $N$ is the total number of episodes, $T^{(n)}$ is the total number of time-steps for episode $n$, $\gamma$ is the discount factor, and the average cumulative importance ratio is $w_t = \frac{1}{N} \sum_{n=1}^{N} \rho_{1:t}^{(n)}$.
	
	\subsubsection{Fitted Q Evaluation (FQE)}
	FQE is a value-based temporal difference algorithm that utilizes the Bellman equation to compute bootstrapped target transitions from collected trajectories and then uses function approximation to compute the Q-value of policy $\pi$. This is formalized as:
	
	\begin{equation}
		V_{FQE}(\pi) = \frac{1}{N} \sum_{n=1}^{N} \sum_{a \in \mathcal{A}} \pi(a|s_1^{(n)}) \hat{Q}_{FQE}^{\pi}(s_1^{(n)}, a)
	\end{equation}
	
	where $\hat{Q}_{FQE}$ is the estimated Q-function of $\pi$.
	
%
%
	
	\section{Results}
	
	\begin{figure}[h!]
		\centering
		\includegraphics[width=0.8\textwidth]{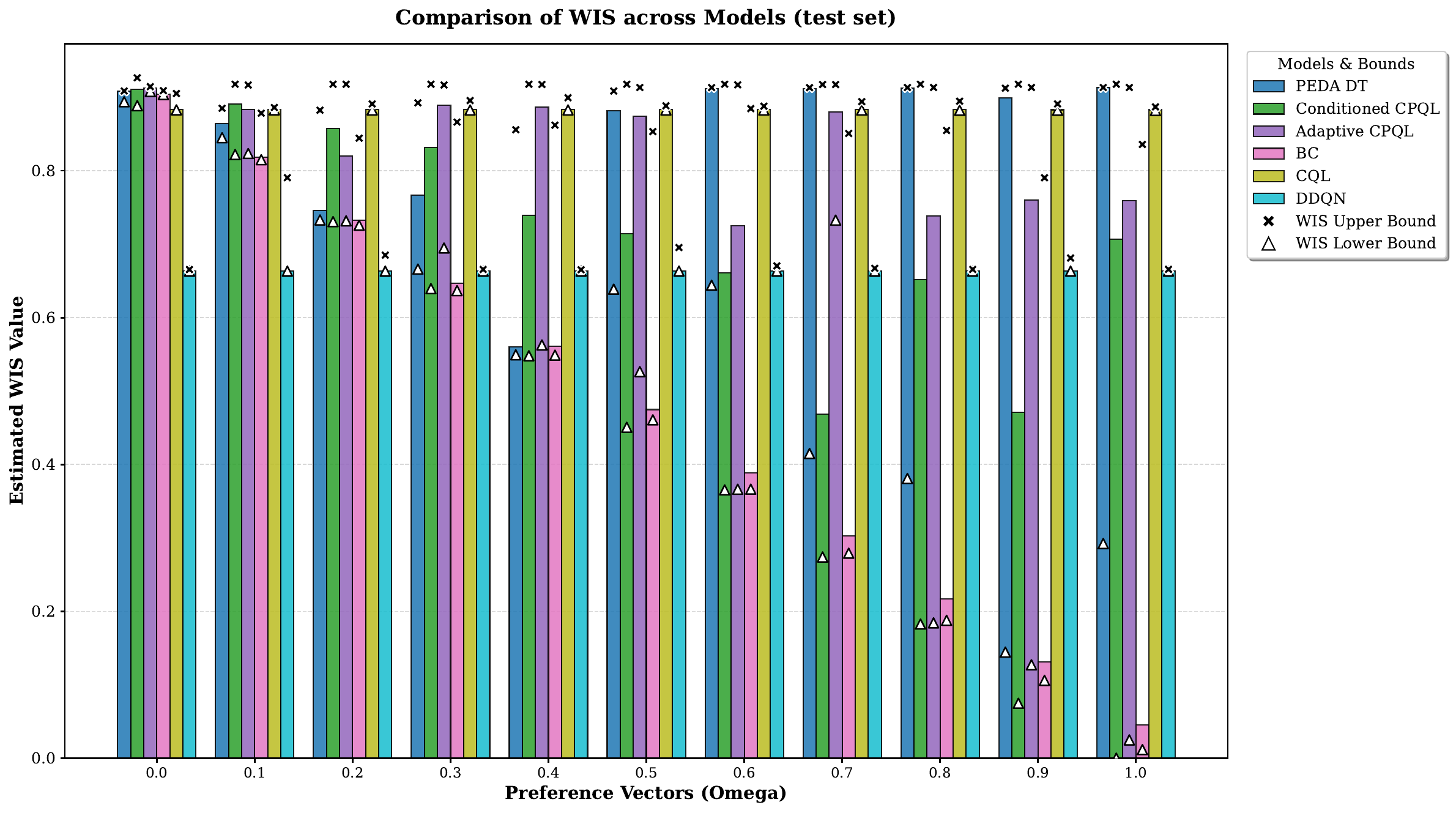}
		\caption{A comparison of different OMORL and OSORL algorithms on the test set based on the WIS metric, where the x-axis represents the preference $x$ given to the mortality reward. This also means that the preference given to length of stay reward is $1-x$.}
		\label{fig:test_wis}
	\end{figure}
	
	\begin{figure}[h!]
		\centering
		\includegraphics[width=0.8\textwidth]{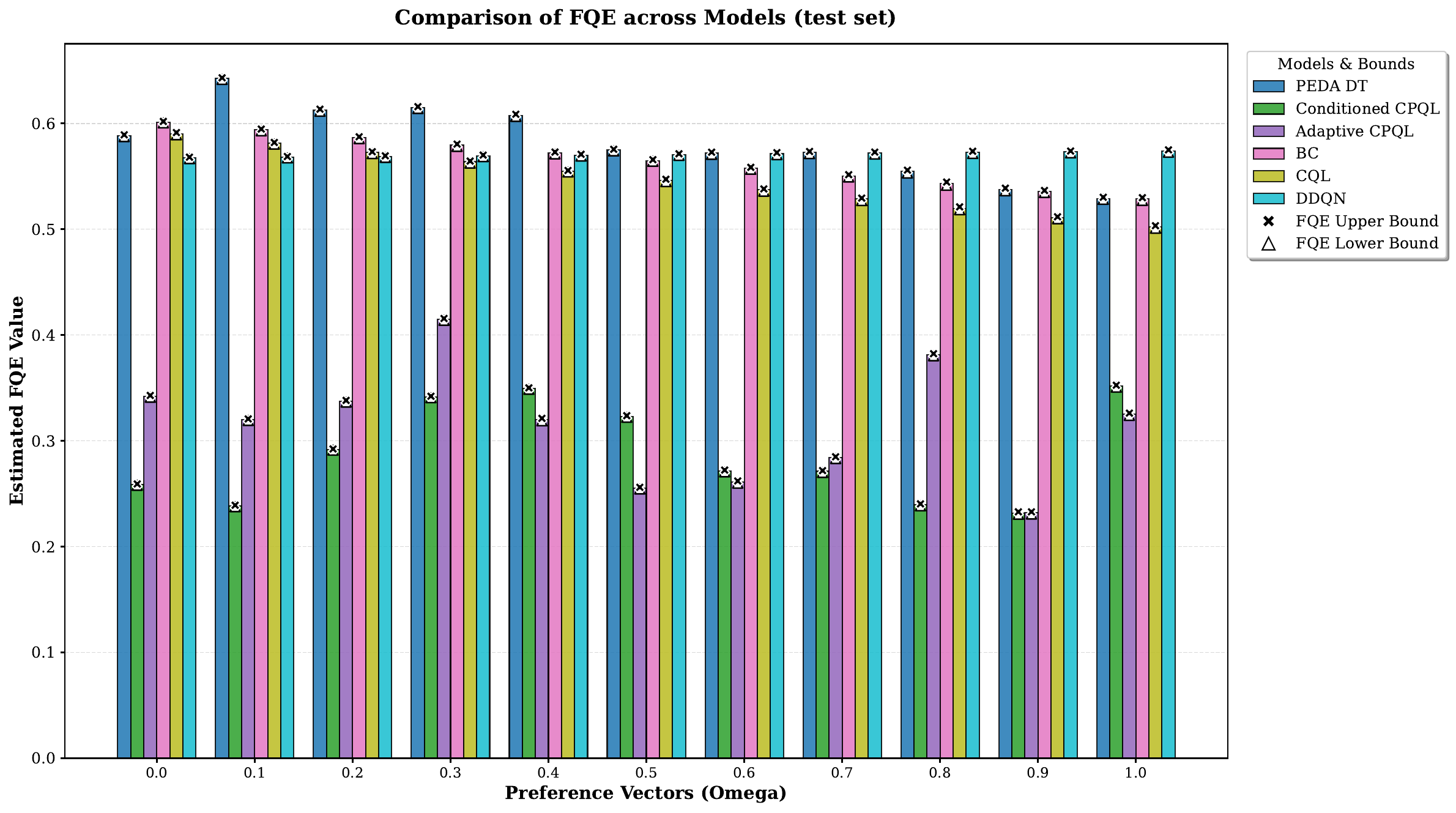}
		\caption{A comparison of different OMORL and OSORL algorithms on the test set based on the FQE metric.}
		\label{fig:test_fqe}
	\end{figure}
	
	
	\newpage
	
	\subsection{OPE Metric Comparison}
	Tables \ref{tab:testwis} and \ref{tab:testfqe} present the estimated values and the width of the Confidence Interval (CI) for WIS and FQE, respectively, across different preference vectors ($\omega$), where CI Width = Upper Bound - Lower Bound.
	
	\begin{table}[h!]
		\footnotesize
		\centering
		\caption{Weighted Importance Sampling (WIS) Scores}
		\label{tab:testwis}
		\begin{tabular}{ccccccc}
			\toprule
			Preferences & PEDA DT & Conditioned CPQL & Adaptive CPQL & BC & CQL & DDQN \\
			\midrule
			$[0.0, 1.0]$ & $0.909\pm0.015$ & $0.911\pm0.038$ & $\mathbf{0.912\pm0.007}$ & $0.904\pm0.005$ & $0.883\pm0.022$ & $0.664\pm0.002$ \\
			$[0.1, 0.9]$ & $0.864\pm0.040$ & $\mathbf{0.891\pm0.096}$ & $0.883\pm0.093$ & $0.819\pm0.063$ & $0.883\pm0.003$ & $0.664\pm0.127$ \\
			$[0.2, 0.8]$ & $0.746\pm0.149$ & $\mathbf{0.857\pm0.187}$ & $0.820\pm0.186$ & $0.733\pm0.119$ & $0.883\pm0.008$ & $0.664\pm0.022$ \\
			$[0.3, 0.7]$ & $0.767\pm0.226$ & $0.831\pm0.278$ & $\mathbf{0.889\pm0.222}$ & $0.647\pm0.229$ & $0.883\pm0.013$ & $0.664\pm0.002$ \\
			$[0.4, 0.6]$ & $0.560\pm0.307$ & $0.740\pm0.370$ & $\mathbf{0.887\pm0.355}$ & $0.561\pm0.313$ & $0.883\pm0.017$ & $0.664\pm0.002$ \\
			$[0.5, 0.5]$ & $\mathbf{0.881\pm0.270}$ & $0.714\pm0.467$ & $0.874\pm0.387$ & $0.475\pm0.393$ & $0.883\pm0.006$ & $0.664\pm0.032$ \\
			$[0.6, 0.4]$ & $\mathbf{0.912\pm0.269}$ & $0.661\pm0.553$ & $0.725\pm0.551$ & $0.389\pm0.518$ & $0.883\pm0.005$ & $0.664\pm0.007$ \\
			$[0.7, 0.3]$ & $\mathbf{0.912\pm0.498}$ & $0.468\pm0.643$ & $0.880\pm0.185$ & $0.303\pm0.572$ & $0.883\pm0.011$ & $0.664\pm0.004$ \\
			$[0.8, 0.2]$ & $\mathbf{0.912\pm0.532}$ & $0.652\pm0.735$ & $0.738\pm0.729$ & $0.217\pm0.667$ & $0.883\pm0.012$ & $0.664\pm0.002$ \\
			$[0.9, 0.1]$ & $\mathbf{0.899\pm0.768}$ & $0.471\pm0.843$ & $0.760\pm0.786$ & $0.131\pm0.685$ & $0.883\pm0.009$ & $0.664\pm0.018$ \\
			$[1.0, 0.0]$ & $\mathbf{0.913\pm0.621}$ & $0.707\pm0.918$ & $0.760\pm0.889$ & $0.045\pm0.824$ & $0.883\pm0.005$ & $0.664\pm0.002$ \\
			\bottomrule
		\end{tabular}
	\end{table}
	
	\begin{table}[h!]
		\footnotesize
		\centering
		\caption{Fitted Q-Evaluation (FQE) Scores}
		\label{tab:testfqe}
		\begin{tabular}{ccccccc}
			\toprule
			Preferences & PEDA DT & Conditioned CPQL & Adaptive CPQL & BC & CQL & DDQN \\
			\midrule
			$[0.0, 1.0]$ & $0.588\pm0.002$ & $0.259\pm0.001$ & $0.342\pm0.002$ & $\mathbf{0.601\pm0.001}$ & $0.590\pm0.002$ & $0.567\pm0.001$ \\
			$[0.1, 0.9]$ & $\mathbf{0.643\pm0.002}$ & $0.238\pm0.001$ & $0.320\pm0.002$ & $0.594\pm0.002$ & $0.581\pm0.002$ & $0.568\pm0.001$ \\
			$[0.2, 0.8]$ & $\mathbf{0.612\pm0.002}$ & $0.292\pm0.001$ & $0.337\pm0.002$ & $0.587\pm0.002$ & $0.572\pm0.002$ & $0.569\pm0.002$ \\
			$[0.3, 0.7]$ & $\mathbf{0.615\pm0.002}$ & $0.341\pm0.001$ & $0.415\pm0.002$ & $0.579\pm0.002$ & $0.564\pm0.002$ & $0.569\pm0.001$ \\
			$[0.4, 0.6]$ & $\mathbf{0.608\pm0.002}$ & $0.349\pm0.001$ & $0.320\pm0.002$ & $0.572\pm0.002$ & $0.555\pm0.001$ & $0.570\pm0.002$ \\
			$[0.5, 0.5]$ & $\mathbf{0.575\pm0.001}$ & $0.323\pm0.002$ & $0.255\pm0.002$ & $0.565\pm0.002$ & $0.546\pm0.002$ & $0.571\pm0.002$ \\
			$[0.6, 0.4]$ & $\mathbf{0.572\pm0.002}$ & $0.271\pm0.002$ & $0.261\pm0.002$ & $0.558\pm0.002$ & $0.537\pm0.002$ & $0.571\pm0.002$ \\
			$[0.7, 0.3]$ & $\mathbf{0.573\pm0.002}$ & $0.271\pm0.002$ & $0.284\pm0.002$ & $0.550\pm0.002$ & $0.528\pm0.002$ & $0.572\pm0.002$ \\
			$[0.8, 0.2]$ & $0.555\pm0.003$ & $0.240\pm0.002$ & $0.381\pm0.002$ & $0.543\pm0.003$ & $0.520\pm0.003$ & $\mathbf{0.572\pm0.002}$ \\
			$[0.9, 0.1]$ & $0.537\pm0.002$ & $0.232\pm0.002$ & $0.232\pm0.002$ & $0.536\pm0.002$ & $0.511\pm0.002$ & $\mathbf{0.573\pm0.002}$ \\
			$[1.0, 0.0]$ & $0.529\pm0.002$ & $0.352\pm0.002$ & $0.325\pm0.002$ & $0.529\pm0.003$ & $0.502\pm0.002$ & $\mathbf{0.574\pm0.002}$ \\
			\bottomrule
		\end{tabular}
	\end{table}

	\subsection{Benchmarking OMORL vs. Scalarized Baselines}
	Our analysis highlights the limitations of scalarized OSORL in dynamic clinical settings. As evidenced in tables \ref{tab:testwis} and \ref{tab:testfqe}, baselines such as BC and CQL generally provide static performance across all preference vectors because they are optimized for a fixed trade-off (typically $\omega = [0.5, 0.5]$). While stable, these models fail to adapt when clinical necessities dictate a shift in priority, such as emphasizing mortality reduction over hospital resources. In contrast, the OMORL algorithms evaluated, specifically PEDA DT, demonstrate the capacity to adjust policy performance based on input preferences, successfully maintaining high estimated returns across varying preferences. Furthermore, by using OMORL algorithms, we avoid the need to train several OSORL models, each for a specific preference vector.
		
	\subsection{Extending Decision Transformers to Multi-Objective Settings}
	A key finding of this benchmark is the robust performance of the Decision Transformer-based approach (PEDA DT) compared to both the Q-learning based MORL variants (CPQL) and standard baselines. Previous work~\citep{rahman2024empowering} demonstrated that single-objective Decision Transformers outperform traditional OSORL algorithms like BC and CQL in sepsis treatment tasks. Our results validate that this performance advantage holds true when extending the architecture to the multi-objective domain. While the prior work established the efficacy of sequence modeling for fixed-reward healthcare tasks, our implementation of PEDA DT confirms that the architecture is uniquely well-suited for the ``stitching'' of suboptimal trajectories required in multi-objective conditioned generation, outperforming the value-based CPQL methods in nearly all tested preference configurations.

	\section{Conclusion and Future Directions}
	
	\subsection{Conclusion}
	
	This paper presented a comprehensive benchmark of OMORL algorithms applied to critical care. By evaluating these models on the MIMIC-IV dataset, we showed that traditional scalarized RL forces a restrictive one-size-fits-all strategy that is often ill-suited for the nuances of patient care. Our experiments indicate that OMORL algorithms, specifically the Modified PEDA Decision Transformer, can successfully learn a diverse range of policies along the Pareto Frontier from offline data. This capability empowers clinicians to weigh the trade-offs between patient survival and resource utilization dynamically at inference time, without the need to retrain models for every new clinical scenario.   
	
	\subsection{Future Directions}
	
	Based on the findings of this benchmark, we identify two critical avenues for future research:
	
	\begin{itemize}
		\item Explainability: While PEDA DT shows high performance, the black-box nature of Transformer models remains a barrier to clinical adoption. Future work should focus on developing explainability methods~\citep{beechey2023explaining} specifically for OMORL, enabling clinicians to understand why a model recommends a specific trade-off between mortality and length of stay.
		\item Model-Based RL: This study focused on model-free approaches. However, recent advancements~\citep{xu2025meddreamer} in SORL have shown that World Models~\citep{ha2018world} can significantly improve sample efficiency and planning capabilities. Extending these model-based approaches to the Multi-Objective setting could provide a more reliable training environment, allowing for the simulation of counterfactual patient trajectories and further improving the robustness of offline policy learning.
	\end{itemize}
	
	\section{Acknowledgements}
	
	We thank Jingwen Ji for processing MIMIC-IV dataset, which was used for training and testing the RL algorithms. We also thank Digital Research Alliance of Canada for the compute resources.
		
	\bibliography{references}
	
	\newpage
	
	\section{Appendix}
	
	\begin{figure}[h!]
		\centering
		\includegraphics[width=0.9\textwidth]{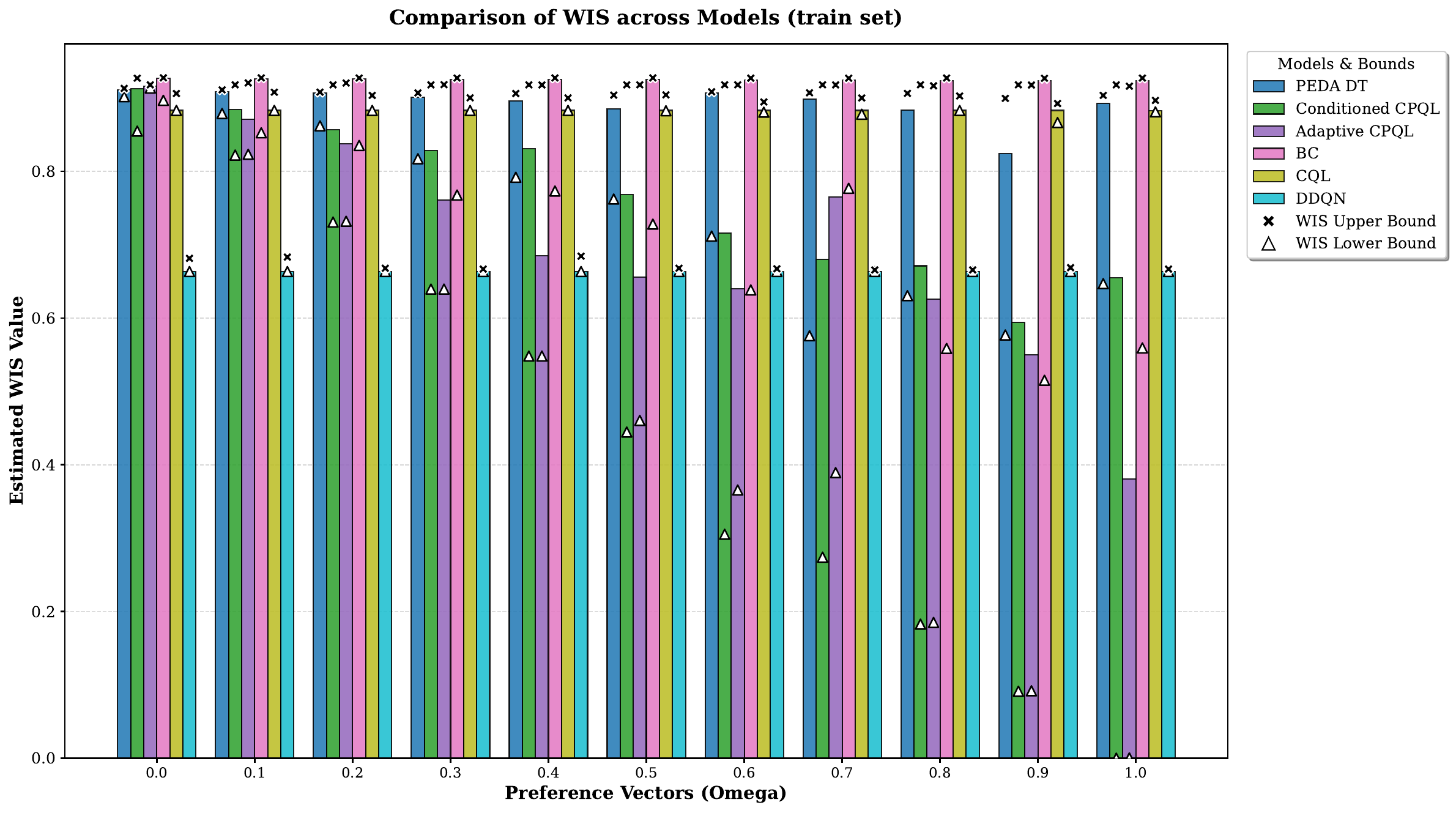}
		\caption{A comparison of different OMORL and OSORL algorithms on the training set based on the WIS metric.}
		\label{fig:train_wis}
	\end{figure}
	
	\begin{figure}[h!]
		\centering
		\includegraphics[width=0.9\textwidth]{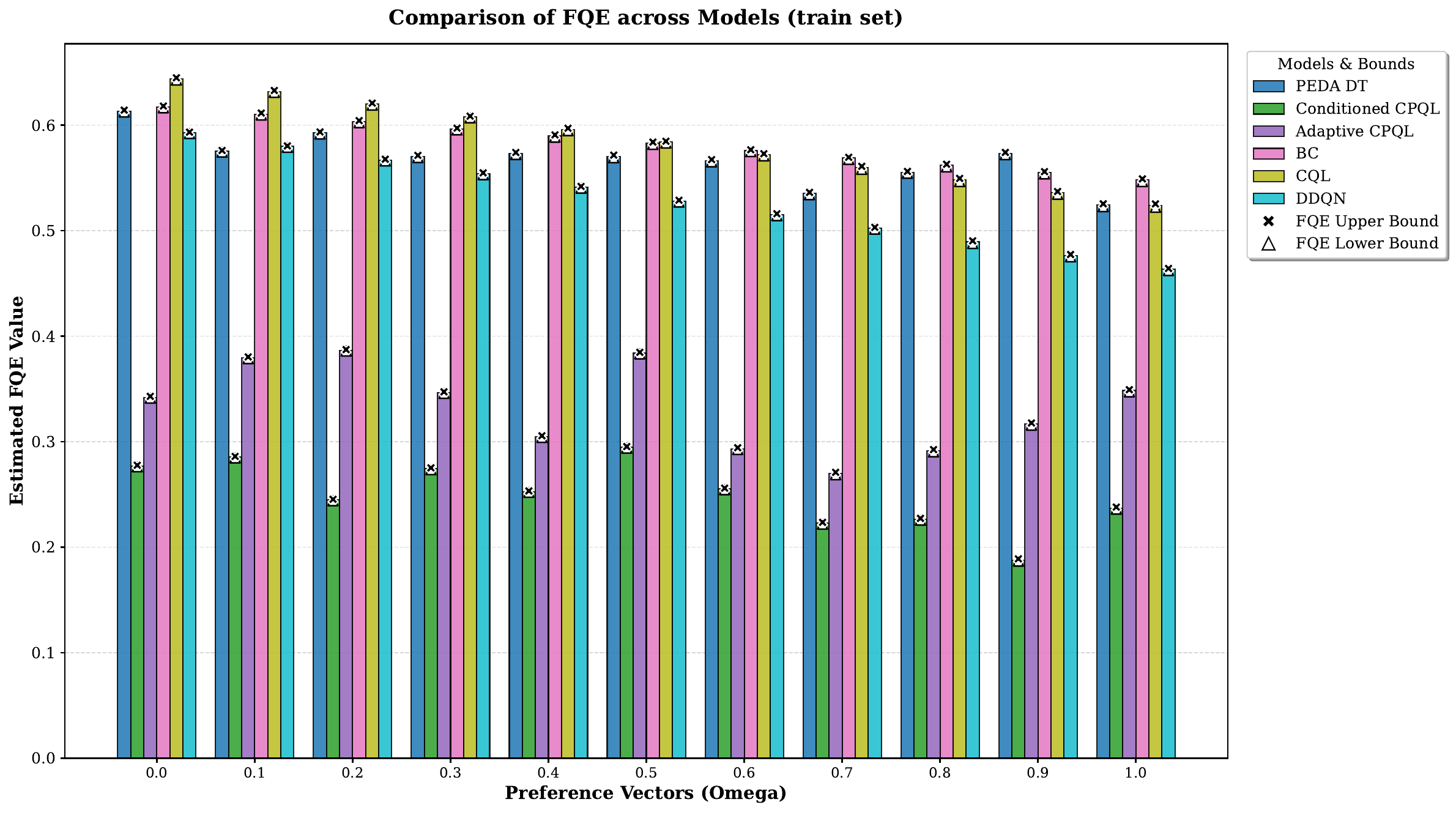}
		\caption{A comparison of different OMORL and OSORL algorithms on the training set based on the FQE metric.}
		\label{fig:train_fqe}
	\end{figure}

	
\end{document}